\documentclass[discussion]{clv3}
\pdfoutput=1
\usepackage[hyphens]{url}
\usepackage{hyperref}
\usepackage{xcolor}
\definecolor{darkblue}{rgb}{0, 0, 0.5}
\hypersetup{colorlinks=true,citecolor=darkblue, linkcolor=darkblue, urlcolor=darkblue}
\usepackage{latexsym}
\usepackage{booktabs}
\usepackage{amsmath}
\usepackage{multirow}
\usepackage[font=normal]{subfig}
\usepackage{adjustbox}
\usepackage[graphicx]{realboxes}
\usepackage{todonotes}
\usepackage{verbatim}
\usepackage{diagbox}

\usepackage{tikz}
\tikzset{
  treenode/.style = {shape=rectangle, rounded corners,
                     draw, align=center,
                     top color=white, bottom color=blue!20},
  root/.style     = {treenode, font=\normalsize, bottom color=red!30},
  env/.style      = {treenode, font=\ttfamily\normalsize},
  dummy/.style    = {circle,draw}
}

\issue{0}{0}{2019}

\runningtitle{Fair is Better than Sensational}

\runningauthor{M. Nissim, R. van Noord and R. van der Goot}

\begin{document}

\title{Fair is Better than Sensational: \\ Man is to Doctor as Woman is to Doctor}

\author{Malvina Nissim}
\affil{University of Groningen}

\author{Rik van Noord}
\affil{University of Groningen}

\author{Rob van der Goot}
\affil{University of Groningen}

\maketitle

\begin{abstract}
Analogies such as \emph{man is to king as woman is to X} are often used to illustrate the amazing power of word embeddings. Concurrently, they have also been used to expose how strongly human biases are encoded in vector spaces built on natural language, like \emph{man is to computer programmer as woman is to homemaker}. 

Recent work has shown that analogies are in fact not such a diagnostic for bias, and other methods have been proven to be more apt to the task. However, beside the intrinsic problems with the analogy task as a bias detection tool, in this paper we show that a series of issues related to how analogies have been implemented and used might have yielded a distorted picture of bias in word embeddings. Human biases \emph{are} present in word embeddings and need to be addressed. Analogies, though, are probably not the right tool to do so. Also, the way they have been most often used has exacerbated some possibly non-existing biases and perhaps hid others. Because they are still widely popular, and some of them have become classics within and outside the NLP community, we deem it important to provide a series of clarifications that should put well-known, and potentially new cases into the right perspective. 
\end{abstract}

\section{Introduction}

Word embeddings are distributed representations of texts which capture similarities between words. Besides improving a wide variety of NLP tasks, the power of word embeddings is often also tested intrinsically.
\citet{mikolov2013efficient} introduced the idea of testing the soundness of embedding spaces via the analogy task.
Analogies are equations of the form $A:B :: C:D$, or simply \textit{A is to B as C is to D}. Given the terms $A,B,C$, the model must return the word that correctly stands for $D$ in the given analogy. A most classic example is \textit{man is to king as woman is to X}, where the model is expected to return \textit{queen}, by subtracting ``manness'' from the concept of king to obtain some general royalty, and then re-adding some ``womanness'' to obtain the concept of queen ($king - man + woman = queen$).

Beside showing this kind of magical power, analogies have been extensively used to show that embeddings carry worrying biases present in our society and thus encoded in language. This bias is often demonstrated by using the analogy task to find stereotypical relations, such as the classic \textit{man is to doctor as woman is to nurse} or \textit{man is to computer programmer as woman is to homemaker}.

The potential of the analogy task has been recently questioned, though. It has been argued that what is observed through the analogy task might be mainly due to irrelevant neighborhood structure rather than to the vector offset that supposedly captures the analogy itself \citep{linzen2016issues,rogers2017too}. Also, \citet{drozd2016word} have shown that the original and classically used \textsc{3cosadd} method \cite{mikolov2013efficient} is not able to capture all linguistic regularities present in the embeddings. And regarding bias, \citet{gonen-goldberg-2019-lipstick} have shown that analogies are not a good diagnostic for bias in embeddings, and have in a way misled bias identification, with consequences on debiasing efforts. 
Indeed, on the recent contextualised embeddings~\citep{peters-etal-2018-deep,devlin-etal-2019-bert}, the analogy task is not used anymore to either evaluate their soundness, or to detect bias \citep{zhao-etal-2019-gender,costa-jussa:19,may-etal-2019-measuring}.

While research indicates that this is the direction that should be pursued to deal with bias in word embeddings, analogies are not only still widely used, but have also left a strong footprint, with some by-now-classic examples often brought up as proof of human bias in language models. A case in point is the opening speech by the ACL President at ACL~2019 in Florence, Italy, where the issue of bias in embeddings is brought up showing biased analogies from a 2019~paper \citep{manzini2019v2}.\footnote{\url{https://www.microsoft.com/en-us/research/uploads/prod/2019/08/ACL-MingZhou-50min-ming.v9-5d5104dcbe73c.pdf}, slide 29.}  

This contribution thus aims at providing some clarifications over the past use of analogies to hopefully raise further and broader awareness of their potential and their limitations, and put well-known and possibly new ones in the right perspective.

First, we take a closer look at the concept of analogy together with requirements and expectations. We look at how the original analogy structure was used to query embeddings, and some misconceptions that a simple implementation choice has caused. More specifically, in the original proportional analogy implementation, all terms of the equation $A:B :: C:D$ are distinct \citep{mikolov2013efficient}. In other words, the model is \textit{forced} to return a \textit{different} concept than any of the original ones. Given an analogy of the form $A:B :: C:D$, the model is not allowed to yield any term $D$ such that $D==B$, $D==A$, or $D==C$, since the  code explicitly prevents this. 
While this constraint is helpful when all terms of the  analogy are expected to be different, it  becomes a problem, and even a dangerous artifact, when the  terms \textit{could} or even \textit{should} be the same.

Second, we discuss different analogy detection strategies/measures that have been proposed, namely the original \textsc{3cosadd} measure, the revised \textsc{3cosmul} measure \citep{levy-goldberg-2014-linguistic}, and the \citet{bolukbasi2016man} formula, which introduces a different take on the analogy construction, reducing the impact of subjective choices.

Third, we highlight the role played by human biases in choosing which analogies to search for, and which results to report. We also show that even when subjective choices are minimised in input (like in \citealt{bolukbasi2016man}), parameter tuning might have consequences on the results, which should not go unnoticed or underestimated.

This work does not mean at all to downplay the presence and danger of human biases in word embeddings. On the contrary: embeddings \textit{do} encode human biases \citep{caliskan2017semantics, garg2018word, kozlowski2018geometry, gonen-goldberg-2019-lipstick}, and we agree that this issue deserves the full attention of the field \citep{hovy-spruit-2016-social}. This is the main reason why transparency over existing and possibly future analogies is crucial.

\section{What counts as \textit{analogy}?}
\label{sec:analogies}

In linguistics, analogies of the form $A:B :: C:D$ can be conceived on two main levels of analysis \cite{fischer:19}. The first one is morphological (so-called strict \textit{proportional analogies}), and they account for systematic language regularities. The second one is more at the lexico-semantic level, and similarities can get looser and more subject to interpretation (e.g. \textit{traffic is to street as water is to riverbed} \cite{turney2012domain}). 
The original, widely used, analogy test set introduced by \citet{mikolov2013efficient} consists indeed of two main categories: morpho-syntactic analogies (\textit{car is to cars as table is to tables}) and semantic analogies (\textit{Paris is to France as Tokyo is to Japan}). Within these, examples are classified in more specific sub-categories.

There are two important aspects that must be considered following the above. First, analogies are (traditionally) mostly conceived as featuring four \textit{distinct} terms. Second, we need to distinguish between cases where there is one specific, expected, correct fourth term, and cases where there isn't. Both aspects bear important methodological consequences in the way we query and analyse (biased) analogies in word embeddings.

\subsection{Should all terms be different?}

Of the four constraints introduced by Turney in formally defining analogies, the last two constraints indirectly force the terms $B$ and $D$ to be different \citep[p.~540]{turney2012domain}. In practice, also all the examples of the original analogy test~\citep{mikolov2013efficient} expect four different terms. Is this always the case? Are expressions featuring the same term twice non-analogies?

Because most out-of-the-box word embeddings have no notion of senses, homographs are modelled as one unit. For example, the infinitive form and the past tense of the verb ``to read'', will be represented by one single vector for the word ``read''. 
A consequence of this is that for certain examples, two terms would be identical, though they would be conceptually different. In strong verbs, infinitive and simple past can be homographs (e.g. split/split), and countries or regions can be homographs with their capitals (e.g. Singapore/Singapore). Other cases where all terms are not necessarily distinct include ``is-a'' relations (hypernyms, \textit{cat:animal :: dog:animal}), and ordered concepts (\textit{silver:gold :: bronze:silver}). 
Moreover, the extended analogy test set created by \citet{gladkova2016analogy} also includes examples where $B$ is the correct answer, for example \textit{country:language} and \textit{thing:color}.
While these examples might not be conceived as standard analogies, the issue with homographs remains. 

\subsection{Is there a correct answer?}

In Mikolov's analogy set, for both macro-categories (morpho-syntactic and semantic), all of the examples are structured such that given the first three terms, there is one specific, correct (expected) fourth term. We can call such analogies ``factual''. While morphosyntactic analogies are in general indeed factual (but there are exceptions due to homographical ambiguities), the picture is rather different for the semantic ones. If we take \textit{man:computer\_programmer :: woman:X} as a semantic analogy, what is the ``correct'' answer? Is there an expected, unbiased completion to this query? Compare it to the case of \textit{he:actor :: she:X} -- it seems quite straightforward to assume that X should be resolved as \textit{actress}. However, such resolution easily rescales the analogy to a morphosyntactic rather than semantic level, thereby also ensuring a factual, unbiased answer. 
 
The morpho-syntactic and semantic levels are indeed not always distinct. When querying \textit{man:computer\_programmer :: woman:X}, or \textit{man:doctor :: woman:X}, is one after a morphosyntactic or a semantic answer? Morpho-syntactically, we should resolve to \textit{doctor}, thereby violating the all-terms-different constraint. If we take the semantic interpretation, there is no single predefined term that ``correctly'' completes the analogy (or maybe \textit{doctor} does here too).\footnote{In this sense, it is admirable that \citet{caliskan2017semantics} try to better understand their results by checking them against actual job distributions between the two genders.}

In such non factual, more creative analogies, various terms could be used for completion depending on the implied underlying relation \cite{turney2012domain}, which could be unclear or unspecified in the query. For the analogies used by \citet{manzini2019v2} (see~Table~\ref{tab:manzini}), for example, it is rather unclear what one would expect to find. Some of the returned terms might be biased, but in order to claim bias, one should also conceive the expected unbiased term. So, if \textit{doctor} is not eligible by violating the distinction constraint, what would the unbiased answer be to the semantic analogy \textit{man:doctor :: woman:X}?

When posing queries, all such aspects should be considered, and one should be aware of what analogy algorithms and implementations are designed to detect. If the correct or unbiased answer to \textit{man:woman :: doctor:X} is expected to be \textit{doctor} and the model is not allowed to return any of the input terms as it would otherwise not abide to the definition of analogy, then such a query should not be asked. If asked anyway under such conditions, the model should not be charged with bias for not returning \textit{doctor}.

\section{Algorithms}
\label{sec:algorithms}
We consider three  strategies that have been used to capture analogies. We use the standard \textsc{3cosadd} function (Eq.~\ref{equ:cosadd}) from \citet{mikolov2013efficient}, and  \textsc{3cosmul}, introduced by~\citet{levy-goldberg-2014-linguistic} to overcome some of the shortcomings of \textsc{3cosadd}, mainly ensuring that a single large term cannot dominate the expression (Eq.~\ref{equ:cosmul}):
\begin{equation}
\underset{d}{\mathrm{argmax}} (\cos (d, c) - \cos (d, a) + \cos (d, b))
\label{equ:cosadd}
\end{equation}
\vspace*{-1cm}
\begin{equation}
\underset{d}{\mathrm{argmax}} \frac{\cos(d, c)\cos(d,b)}{\cos(d, a) + 0.001}
\label{equ:cosmul}
\end{equation}
\citet{bolukbasi2016man} designed another formula, specifically focused on finding pairs $B:D$ with a similar direction as $A:C$:

\begin{equation}
S_{(a,c)} (b,d) = 
\begin{cases}
\cos(a-c, b-d) & \mbox{if} ||b-d|| \leq \delta\\
0   & otherwise \\
\end{cases}
\label{equ:bol}
\end{equation}
\noindent They do not assume that $B$ is known beforehand, and generate a ranked list of $B:D$ pairs, with the advantage of introducing less subjective bias in the input query (see Section~\ref{sec:subjective}). To ensure that $B$ and $D$ are related, the threshold $\delta$ is introduced, and set to 1.0 in~\citet{bolukbasi2016man}. This corresponds to $\pi/3$ and in practice means that $B$ and $D$ have to be closer together than two random embedding vectors.
If $B$ is chosen beforehand (as in classic analogies), it is more efficient to transform the formula to only search for $D$: 

\begin{equation}
\underset{d}{\mathrm{argmax}} 
\begin{cases}
\cos(a-c, b-d) & \mbox{if} ||b-d|| \leq \delta\\
0   & otherwise \\
\end{cases}
\label{equ:bol2}
\end{equation}

\noindent Note that the resulting scores for the same analogy are exactly the same as in Equation~\ref{equ:bol}. This formula allows one to examine the top-N $B:D$ candidates for a specific query, while Equation~\ref{equ:bol} will only return one single $B:D$ pair for a given $B$.

Even though it is not part of the equations, in practice most implementations of these optimization functions specifically ignore one or more input vectors. Most likely, this is because the traditional definitions of analogy require all terms to be different (see Section~\ref{sec:analogies}), and the original analogy test set reflects this. However, we have seen that this is a strong constraint, both in  morphosyntactic and semantic analogies. 
Moreover, even though this constraint is mentioned in the original paper \citep{mikolov2013efficient} and in follow-up work \citep{linzen2016issues, bolukbasi2016man, rogers2017too, goldberg2017neural,schluter-2018-word}, we believe this is not common knowledge in the field (analogy examples are still widely used), and even more so outside the field.\footnote{This was confirmed by the response we got when we uploaded a first version of the paper.}

\section{Is the bias in the models, in the implementation, or in the queries?}

In addition to preventing input vectors from being returned, other types of implementation choices (such as punctuation, capitalization, or word frequency cutoffs), and subjective decisions play a substantial role. 
So, what is the actual influence of such choices on obtaining biased responses? 

In what follows, unless otherwise specified, we run all queries on the standard GoogleNews embeddings.\footnote{\url{https://code.google.com/archive/p/word2vec/}}
All code to reproduce our experiments is available: \url{https://bitbucket.org/robvanderg/w2v}.

\subsection{Ignoring or allowing the input words}
In the default implementation of word2vec~\citep{mikolov2013efficient}, gensim~\citep{rehurek_lrec} as well as the code from~\citet{bolukbasi2016man}, the input terms of the analogy query are not allowed to be returned.\footnote{ In Equation~\ref{equ:bol}, in practice B will almost never be returned, as it will always be assigned a score of 0.0, making it the last ranked candidate.}
We adapted all these code-bases to allow for the input words to be returned.\footnote{The \textsc{3cosadd} unconstrained setting can be tested in an online demo: \url{www.robvandergoot.com/embs}} 

To compare the different methods, we evaluated all of them on the test set from~\citet{mikolov2013efficient}. The results  in Table~\ref{tab:miko} show a large drop in performance for \textsc{3cosadd} and \textsc{3cosmul} in the unconstrained setting. In most cases, this is because the second term is returned as answer (\textit{man is to king as woman is to \underline{\smash{king}}}, thus $D==B$),  but in some cases it is the  third term  that gets returned (\textit{short is to shorter as new is to \underline{new}}, thus $D==C$). A similar drop in performance was observed before by \citet{linzen2016issues} and \citet{schluter-2018-word}.
The \citet{bolukbasi2016man} method shows very low scores, but this was to be expected, since their formula was not specifically designed to capture factual analogies. But what is so different between factual and biased analogies? 

\begin{table}[!htb]
    \centering
    \setlength{\tabcolsep}{4pt}
    \resizebox{\textwidth}{!}{
    \begin{tabular}{l | c c | c c | c c}
        \toprule
                 & \bf \textsc{3cosadd} & \bf uncon. & \bf \textsc{3cosmul} & \bf uncon. & \bf \textsc{bolukbasi} & \bf uncon. \\
        \midrule
       Google Analogy - micro  & 0.74                 & 0.21       & 0.75                 & 0.47       & 0.04           & 0.11      \\
       Google Analogy - macro  & 0.71                 & 0.21       & 0.73                 & 0.45       & 0.06           & 0.11       \\
        \bottomrule
    \end{tabular}

    }
    \caption{Accuracies of the three formulas on the Google Analogy test set \citep{mikolov2013efficient}, comparing the constrained version (original code), with the unconstrained version (uncon.). For all formulas, unconstrained means also taking the input vectors into account. For \textsc{bolukbasi}, more constraints were removed (see Section~\ref{sec:constraints}). }
    \label{tab:miko}
\end{table}

\begin{table}[!htb]
    \setlength\tabcolsep{4pt}
    \centering
    \resizebox{\textwidth}{!}{
    \begin{tabular}{l | l | l | l}
        \toprule
            & \multicolumn{1}{l|}{
                \begin{tabular}{l l l }
                    man & : & doctor\\
                    woman & : & \multicolumn{1}{c}{X} \\
                \end{tabular}
            }
            & \multicolumn{1}{l|}{
                \begin{tabular}{l l l }
                    he & : & doctor\\
                    she & : & \multicolumn{1}{c}{X} \\
                \end{tabular}
            }
            & \multicolumn{1}{l}{
                \begin{tabular}{l l l }
                    man & : & computer\_programmer\\
                    woman & : & \multicolumn{1}{c}{X} \\
                \end{tabular}
            }\\
        \midrule
        \textsc{3cosadd} & \hspace{.125cm} gynecologist & \hspace{.125cm} nurse & \hspace{.125cm} homemaker \\
        \quad unconstrained & \hspace{.125cm} doctor & \hspace{.125cm} doctor & \hspace{.125cm} computer\_programmer \\
        \midrule
        \textsc{3cosmul} & \hspace{.125cm} gynecologist & \hspace{.125cm} nurse & \hspace{.125cm} homemaker \\
        \quad unconstrained & \hspace{.125cm} doctor & \hspace{.125cm} doctor & \hspace{.125cm} computer\_programmer \\
        \midrule
       \textsc{bolukbasi} & \hspace{.125cm} midwife & \hspace{.125cm} nurse & \hspace{.125cm}  \\
        \quad unconstrained & \hspace{.125cm} gynecologist & \hspace{.125cm} nurse & \hspace{.125cm} schoolteacher \\
        \bottomrule
    \end{tabular}
    }
    \caption{Example output of the three algorithms for their regular and unconstrained implementations for three well-known gender bias analogies.}
    \label{tab:examples2}
\end{table}

In Table~\ref{tab:examples2}, we report the results using the same settings for  a small selection of mainstream examples from the literature on embedding bias. It directly becomes clear that removing constraints leads to  different (and arguably less biased) results.\footnote{This was noticed before: \url{https://medium.com/artists-and-machine-intelligence/ami-residency-part-1-exploring-word-space-andprojecting-meaning-onto-noise-98af7252f749} and \url{https://www.youtube.com/watch?v=25nC0n9ERq4}} More precisely, for \textsc{3cosadd} and \textsc{3cosmul} we get word $B$ as answer, and using the method described by~\citet{bolukbasi2016man} we get different results because with the vocabulary cutoff they used (50,000 most frequent words, see Section~\ref{sec:constraints}), \textit{gynecologist} (51,839) and \textit{computer\_programmer} (57,255) were excluded.\footnote{Though \textit{man is to computer programmer as woman is to homemaker} is used in the title of \citet{bolukbasi2016man}, this analogy is obtained using \textsc{3cosadd}.} 

The analogy \textit{man is to doctor as woman is to nurse} in Table~\ref{tab:examples2} is a classic showcase of human bias in word embeddings. This  biased analogy reflecting gendered stereotypes in our society, is however truly meaningful  only if the system were allowed to yield \textit{doctor} (arguably the expected answer in absence of bias, see Section~\ref{sec:analogies}) instead of \textit{nurse}, and it doesn't. But using the original analogy code it is impossible to obtain \textit{man is to doctor as woman is to doctor} (where $D==B$). Under such commonly used settings, it is not exactly fair to claim the embedding space is biased because it does not return \textit{doctor}.

\subsection{Subjective factors}
\label{sec:subjective}

Let us take a step back though, and ask: Why do people query \textit{man is to doctor as woman is to X?} 
In fairness, one should wonder how much bias leaks in from our own views, preconceptions, and expectations. In this section we aim to show how these affect the queries we pose and the results we get, and how the inferences we can draw depend strongly on the choices we make in formulating queries and in reporting the outcome.

To start with, the large majority of the queries we pose and find in the literature imply human bias. People usually query for \textit{man:doctor :: woman:X}, which in \textsc{3cosadd} and \textsc{3cosmul} is different than querying for \textit{woman:doctor :: man:X}, both in results and in assumptions (often expecting to find a biased answer). This issue also raises the major, usually unaddressed question as to what would the unbiased, desired, $D$ term be? Such bias-searching queries do not pose factual, one-correct-answer, analogies, unless interpreted morpho-syntactically (See Section~\ref{sec:analogies}).

\begin{table}[!b]
    \centering
    \setlength{\tabcolsep}{4pt}
    \renewcommand{\arraystretch}{1.1}
    \resizebox{\textwidth}{!}{
    \begin{tabular}{l l l r l}
        \toprule
        & \textbf{Analogy}              & \textbf{Reported} & \textbf{Idx} & \textbf{Top-5 answers (averaged)}       \\
        \midrule
        \multirow{6}{*}{\rotatebox[origin=c]{90}{\footnotesize{\citet{manzini2019v2}}}} & caucasian lawful black      & criminal          & 2.0       & lawful criminal defamation libel vigilante  \\
        & asian yuppie caucasian      & hillbilly         & 5.0       & yuppie knighting pasty hipster hillbilly              \\
        & black killer asian          & engineer          & 5.2       & addict aspie impostor killer engineer                 \\
        & jew liberal christian       & conservative      & 2.0       & liberal conservative progressive heterodox secular    \\
        & jew journalist muslim       & terrorist         & 1.6       & terrorist purportedly journalist watchdog cia         \\
        & muslim regressive christian & conservative      & 9.2       & regressive progressive milquetoast liberal neoliberal \\
         \midrule
       \multirow{6}{*}{\rotatebox[origin=c]{90}{\footnotesize{\citet{manzini2019v3}}}} & black homeless caucasian    & servicemen        & 211.6      & homeless somalis unemployed bangladeshi nigerians \\
        & caucasian hillbilly asian   & suburban          & 60.6       & hillbilly hippy hick redneck hippie               \\
        & asian laborer black         & landowner         & 3.0        & laborer landowner fugitive worker millionaire     \\
        & jew greedy muslim           & powerless         & 8.8        & greedy corrupt rich marginalized complacent       \\
        & christian familial muslim   & warzone           & 7172       & familial domestic marital bilateral mutual        \\
        & muslim uneducated christian & intellectually    & 16.6       & uneducated uninformed idealistic elitist arrogant \\
        \bottomrule
    \end{tabular}
    }
    \caption{Overview of reported biased analogies in \citet{manzini2019v2} and \citet{manzini2019v3}, obtained using the \textsc{3cosadd} method without constraints, but their embeddings as they are (with constraints on the vocabulary). ``Idx'' refers to the average position of the reported biased word as we find it in their five embedding sets (trained on Reddit data) trained with different seeds (i.e., the same space they used).}
    \label{tab:manzini}
\end{table}

Another subjective decision has to do with reporting results. One would think that the top returned term should always be reported, or possibly the top five, if willing to provide a broader picture. However, subjective biases and result expectation might lead to discard returned terms that are not viewed as biased, and report biased terms that are however appearing further down in the list. This causes a degree of arbitrariness in reporting results that can be substantially misleading. 
As a case in point, we discuss here the recent ``Manzini et al.'' paper, which is the work from which the examples used in the opening presidential speech of ACL~2019 were taken (see Footnote~1). This paper was published in three subsequent versions, differing only in the analogy queries used and the results reported. We discuss this to show how subjective the types of choices above can be, and that unless very transparent about methodology and implementation, one can almost claim anything they desire.

In the first version of their paper \citep{manzini2019naacl}, the authors accidentally searched for the inverse of the intended query: instead of asking the model \textit{A is to B as C is to X} (\textit{black is to criminal as caucasian is to X}), they queried \textit{C is to B as A is to X} (\textit{caucasian is to criminal as black is to X}).\footnote{We confirmed this with the authors.}  What is surprising is that they still managed to find biased examples by inspecting the top-N returned $D$~terms. 
In other words, they reported the analogy \textit{black is to criminal as caucasian is to police} to support the hypothesis that there is cultural bias against the black, but the analogy they had in fact found was \textit{caucasian is to criminal as black is to police}, so the complete opposite.
This should make us extremely wary of how easy it can be to find biased analogies when specifically looking for them. 

They fixed this mistake in their second and third version \citep{manzini2019v2, manzini2019v3}. However, it is unclear from the text which algorithm is used to obtain these analogies. We tried the three algorithms described in Section~\ref{sec:algorithms}, and in Table~\ref{tab:manzini} we show the results of \textsc{3cosadd}, for which we could most closely reproduce their results (for both versions).

For their second version, in 5 out of their 6 examples the input word $B$ would actually be returned before the reported answer $D$. For three of the six analogies, they pick a term from the returned top-10 rather than the top returned one. In their third version \citep{manzini2019v3}, the authors change the list of tested analogies, especially regarding the $B$ terms. It is unclear under which assumption some of these `new' terms were chosen to be tested (\textit{greedy} associated to \textit{jew}, for example: what is one expecting to get -- biased or non-biased -- considering this is a negative stereotype to start with, and the $C$ term is \textit{muslim}?). However, for each of the analogy algorithms, we cannot reasonably reproduce four out of six analogies, even when inspecting the top~10 results. 

While qualitatively observing and weighing the bias of a large set of returned answers can make sense, it can be misleading to cherry-pick and report very biased terms in sensitive analogies. At the very least, when reporting term-N, one should  report the top-N terms to provide a more complete picture.  For example, in the top-10 results for \textit{man is to doctor as woman is to X} using \textsc{3cosadd}, some terms refer to medical professions that have only women as patients (\textit{gynecologist}, \textit{obstetrician}, \textit{ob\_gyn}, \textit{midwife}), going to show that it is not always clear which semantic relation is implied in the queries (in this case: professions or patients?), and more than one can be present and confounded. After all, if we query the Google embeddings using unconstrained \textsc{3cosadd} for \textit{man (or woman) is to doctor as dog (or animal) is to X}, we get \textit{veterinarian}.

\subsection{Other constraints}
\label{sec:constraints}

Using the \textsc{bolukbasi} formula is much less prone to subjective choices. It takes as input only two terms ($A$ and $C$, like \textit{man} and \textit{woman}), thus reducing the bias present in the query itself, and consequently the impact of human-induced bias expectation. At the same time, though, starting with $A:C$, the formula requires some parameter tuning in order to obtain (a) meaningful $B:D$ pair(s). Parameter values, together with other pre-processing choices, also affect the outcome, possibly substantially, and must be weighed in when assessing bias.

As shown in Eq.~\ref{equ:bol}, \citet{bolukbasi2016man} introduce a threshold $\delta$ to ensure that $B$ and $D$ are semantically similar. In their work, $\delta$ is set to 1 to ensure that $B$ and $D$ are closer than two random vectors (see Section~\ref{sec:algorithms}). Choosing alternative values for $\delta$ will however yield quite different results, and it is not a straightforward parameter to tune, since it cannot be done against some gold standard, ``correct'' examples. 

\begin{table}[!htb]
    \centering
    \resizebox{\textwidth}{!}{

    \begin{tabular}{l l l l l l}
        \toprule
                & \multicolumn{4}{c}{\textbf{Threshold ($\delta$)}} \\
        \textbf{Voc. size} &  \bf 0.8          & \bf 0.9          & \bf 1.0          & \bf 1.1            & \bf 1.2            \\
        \midrule
        10,000      & doctors      & nurse        & nurse        & nurse          & woman          \\
        25,000      & doctors      & nurse        & nurse        & nurse          & woman          \\
        50,000      & doctors      & nurse        & \textit{midwife}      & midwife        & woman          \\
        100,000     & gynecologist & gynecologist & gynecologist & gynecologist   & gynecologist   \\
        250,000     & gynecologist & gynecologist & gynecologist & gynecologist   & gynecologist   \\
        500,000     & gynecologist & gynecologist & gynecologist & nurse\_midwife & nurse\_midwife \\
        full vocab.& gynecologist & gynecologist & gynecologist & nurse\_midwife & nurse\_midwife \\
        \bottomrule
    \end{tabular}}
    \caption{Influence of vocabulary size and threshold value for the method of \citet{bolukbasi2016man}. With extreme values for the threshold, and allowing to return query words, the answer becomes ``doctor'' ($\leq$0.5) and ``she'' ($\geq$1.5).  Italics: original settings.}
    \label{tab:tuning}
    \vspace{-0.55cm}
\end{table}

Another common constraint that can have a substantial impact on the results is limiting the embedding set to the top-N most frequent words. 
Both \citet{bolukbasi2016man} and~\citet{manzini2019naacl} filter the embeddings to only the 50,000 most frequent words, though no motivation for this need or this specific value is provided. Setting such an arbitrary value might result in the exclusion of valid alternatives. Further processing can also rule out potentially valid strings. For example, \citet{manzini2019naacl} lowercase all words before training, and remove words containing punctuation after training, whereas~\citet{bolukbasi2016man} keep only words that are shorter than 20 characters and do not contain punctuation or capital letters (after training the embeddings).  

In order to briefly illustrate the impact of varying the values of the $\delta$ threshold and the vocabulary size when using the \textsc{bolukbasi} formula, in Table~\ref{tab:tuning} we show the results when changing them for the query \textit{man is to doctor as woman is to X}.\footnote{Given that $B$ is known, we use the formula in Eq.~\ref{equ:bol2}.} The variety of answers, ranging from what can be considered to be biased (\textit{nurse}) to not biased at all (\textit{doctors}), illustrates how important it is to be aware of the influence of choices concerning  implementation and parameter values.

\section{Final remarks}

If analogies might not be the most appropriate tool to capture certain relations, surely matters have been made worse by the way that consciously or not they have been used.  \citet{gonen-goldberg-2019-lipstick} have rightly dubbed them sensational ``party tricks'', and this is harmful for at least two reasons. One is that they get easily propagated both in science itself \cite{jha2017does,gebru2018datasheets,mohammad2018semeval, hall-maudslay-etal-2019-name}, also outside NLP and AI \cite{mcquillan2018people} and in popularised articles \cite{zou2018ai}, where readers are usually in no position to verify the reliability or significance of such examples. The other is that they might mislead the search for bias and the application of debiasing strategies. And while it is debatable whether we should aim at removal or rather at transparency and awareness \cite{caliskan2017semantics,gonen-goldberg-2019-lipstick}, it is crucial that we are clear and transparent about what analogies can and cannot do as a diagnostic for embeddings bias, and about all the implications of subjective and implementation choices.
This is a strict pre-requisite to truly understand how and to what extent embeddings  encode and reflect the biases of our society, and  how to cope with this, both socially and computationally.
\\

\clearpage
\starttwocolumn
\bibliography{compling_style}

\end{document}